\documentclass[11pt,a4paper]{article}
\usepackage[hyperref]{emnlp-ijcnlp-2019}
\usepackage{times}
\usepackage{latexsym}
\usepackage{booktabs}
\usepackage{amsfonts}
\usepackage{color}
\usepackage{graphicx}
\usepackage{amsmath}
\usepackage{amssymb}

\usepackage{array}
\newcolumntype{L}[1]{>{\raggedright\let\newline\\\arraybackslash\hspace{0pt}}m{#1}}
\newcolumntype{C}[1]{>{\centering\let\newline\\\arraybackslash\hspace{0pt}}m{#1}}
\newcolumntype{R}[1]{>{\raggedleft\let\newline\\\arraybackslash\hspace{0pt}}m{#1}}
\usepackage{dblfloatfix}

\usepackage{url}

\aclfinalcopy 

\title{Clause-Wise and Recursive Decoding for Complex and Cross-Domain Text-to-SQL Generation}

\author{Dongjun Lee \\
  SAP Labs Korea \\
  {\tt dongjun.lee01@sap.com} \\}

\date{}

\begin{document}
\maketitle
\begin{abstract}
Most deep learning approaches for text-to-SQL generation are limited to the WikiSQL dataset, which only supports very simple queries over a single table. We focus on the \textit{Spider} dataset, a complex and cross-domain text-to-SQL task, which includes complex queries over multiple tables.
In this paper, we propose a SQL clause-wise decoding neural architecture with a self-attention based database schema encoder to address the \textit{Spider} task. Each of the clause-specific decoders consists of a set of sub-modules, which is defined by the syntax of each clause. Additionally, our model works recursively to support nested queries.
When evaluated on the \textit{Spider} dataset, our approach achieves 4.6\% and 9.8\% accuracy gain in the test and dev sets, respectively. In addition, we show that our model is significantly more effective at predicting complex and nested queries than previous work.
\end{abstract}
\section{Introduction}
Text-to-SQL generation is the task of translating a natural language question into the corresponding SQL. Recently, various deep learning approaches have been proposed based on the WikiSQL dataset \citep{wikisql}. However, because WikiSQL contains only very simple queries over just a single table, these approaches \citep{sqlnet, meta-learning, typesql, coarse-to-fine} cannot be applied directly to generate complex queries containing elements such as \texttt{JOIN}, \texttt{GROUP BY}, and nested queries.

To overcome this limitation, \citet{spider} introduced \textit{Spider}, a new complex and cross-domain text-to-SQL dataset. It contains a large number of complex queries over different databases with multiple tables. It also requires a model to generalize to unseen database schema as different databases are used for training and testing. Therefore, a model should understand not only the natural language question but also the schema of the corresponding database to predict the correct SQL query.
 
In this paper, we propose a novel SQL-specific clause-wise decoding neural network model to address the \textit{Spider} task. We first predict a sketch for each SQL clause (e.g., \texttt{SELECT}, \texttt{WHERE}) with text classification modules. Then, clause-specific decoders find the columns and corresponding operators based on the sketches. Our contributions are summarized as follows.
\begin{itemize}
    \item We decompose the clause-wise SQL decoding process. We also modularize each of the clause-specific decoders into sub-modules based on the syntax of each clause. Our architecture enables the model to learn clause-dependent context and also ensures the syntactic correctness of the predicted SQL.
    \item Our model works recursively so that it can predict nested queries.
    \item We also introduce a self-attention based database schema encoder that enables our model to generalize to unseen databases.
\end{itemize}

In the experiment on the \textit{Spider} dataset, we achieve 24.3\% and 28.8\% exact SQL matching accuracy on the test and dev set respectively, which outperforms the previous state-of-the-art approach \citep{syntaxsqlnet} by 4.6\% and 9.8\%. In addition, we show that our approach is significantly more effective compared to previous work at predicting not only simple SQL queries, but also complex and nested queries.

\section{Related Work}
Our work is related to the grammar-based constrained decoding approaches for semantic parsing \citep{yin2017syntactic, ASN, iyer2018mapping}. While their approaches are focused on general purpose code generation, we instead focus on SQL-specific grammar to address the text-to-SQL task. Our task differs from code generation in two aspects. First, it takes a database schema as an input in addition to natural language. To predict SQL correctly, a model should fully understand the relationship between the question and the schema. Second, as SQL is a non-procedural language, predictions of SQL clauses do not need to be done sequentially.

For text-to-SQL generation, several SQL-specific approaches have been proposed \citep{wikisql, sqlnet, meta-learning, typesql, coarse-to-fine, yavuz2018takes} based on WikiSQL dataset \citep{wikisql}. However, all of them are limited to the specific WikiSQL SQL sketch, which only supports very simple queries. It includes only the \texttt{SELECT} and \texttt{WHERE} clauses, only a single expression in the \texttt{SELECT} clause, and works only for a single table.
To predict more complex SQL queries, sequence-to-sequence \citep{iyer, improve-text-to-sql} and template-based \citep{improve-text-to-sql, lee2019one} approaches have been proposed. However, they focused only on specific databases such as ATIS \cite{atis} and GeoQuery \cite{geo}. Because they only considered question and SQL pairs without requiring an understanding of database schema, their approaches cannot generalize to unseen databases. 

SyntaxSQLNet \citep{syntaxsqlnet} is the first and state-of-the-art model for the \textit{Spider} \citep{spider}, a complex and cross-domain text-to-SQL task. They proposed an SQL specific syntax tree-based decoder with SQL generation history.
Our approach differs from their model in the following aspects.
First, taking into account that SQL corresponds to non-procedural language, we develop a clause-specific decoder for each SQL clause, where SyntaxSQLNet predicts SQL tokens sequentially.
For example, in SyntaxSQLNet, a single column prediction module works both in the \texttt{SELECT} and \texttt{WHERE} clauses, depending on the SQL decoding history. In contrast, we define and train decoding modules separately for each SQL clause to fully utilize clause-dependent context.
Second, we apply sequence-to-sequence architecture to predict columns instead of using the sequence-to-set framework from SyntaxSQLNet, because correct ordering is essential for the \texttt{GROUP BY} and \texttt{ORDER BY} clauses.
Finally, we introduce a self-attention mechanism \citep{self-attention} to efficiently encode database schema, which includes multiple tables.

\begin{figure}[t]
	\centering\includegraphics[scale=0.31]{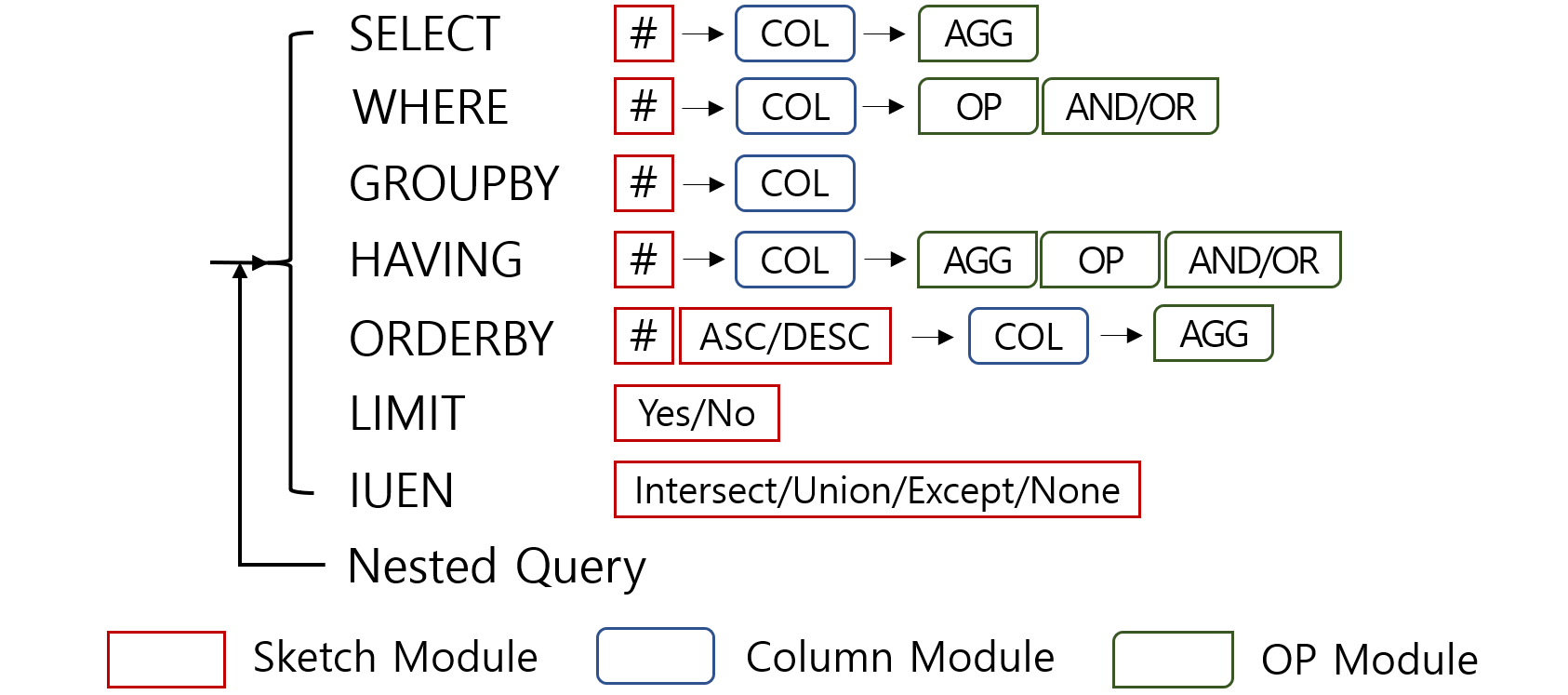}
	\caption{\label{figure1} Clause-wise and recursive SQL generation process.}
\end{figure}

\section{Methodology}
We predict complex SQL clause-wisely as described in Figure~\ref{figure1}. Each clause is predicted consecutively by at most three different types of modules (sketch, column, operator). The same architecture recursively predicts nested queries with temporal predicted SQL as an additional input.

\subsection{Question and Schema Encoding}
\label{sec:encoding}
We encode a natural language question with a bi-directional LSTM. We denote $H_{Q} \in \mathbb{R}^{d \times \left\vert X \right\vert}$ as the question encoding, where $d$ is the number of LSTM units and $\left\vert X \right\vert$ is the number of tokens in the question. 

To encode a database schema, we consider each column in its tables as a concatenated sequence of words from the table name and column name with a separation token. (e.g., [student, \texttt{[SEP]}, first, name]). First, we apply bi-directional LSTM over this sequence for each column. Then, we apply the self-attention mechanism \citep{self-attention} over the LSTM outputs to form a summarized fixed-size vector for each column. For the $i$th column, its encoding $h_{col}^{(i)} \in \mathbb{R}^{d}$ is computed by a weighted sum of the LSTM output $o_{col}^{(i)} \in \mathbb{R}^{d \times \left\vert L \right\vert}$ as follows:
\begin{gather}
\alpha = \texttt{softmax}(w^T \texttt{tanh}(o_{col}^{(i)})) \\
h_{col}^{(i)} = o_{col}^{(i)} \, \alpha^T
\end{gather}
where $\left\vert L \right\vert$ is the number of tokens in the column and $w \in \mathbb{R}^{d}$ is a trainable parameter.
We denote $H_{col} = [h_{col}^{(1)}, ... h_{col}^{(\left\vert C \right\vert)}]$ as columns encoding where $\left\vert C \right\vert$ is the number of columns in the database.


\subsection{Sketch Prediction}
\label{sec:sketch}
We predict the clause-wise sketch via 8 different text classification modules that include the number of SQL expressions in each clause, the presence of \texttt{LIMIT} clause, and the presence of \texttt{INTERSECT/UNION/EXCEPT} as described in Figure~\ref{figure1}.
All of them share the same model architecture but are trained separately. For the classification, we applied attention-based bi-directional LSTM following \citet{attention-text-clf}.

First, we compute sentence representation $r_s \in \mathbb{R}^{d}$ by a weighted sum of question encoding $H_{Q} \in \mathbb{R}^{d \times \left\vert X \right\vert}$. Then we apply the softmax classifier to choose the sketch as follows:
\begin{gather}
\alpha_{s} = \texttt{softmax}(w_{s}^T \texttt{tanh}(H_{Q})) \\
r_s = H_{Q} \, \alpha_{s}^T \\
P_{sketch} = \texttt{softmax}(W_{s}r_s + b_{s})
\end{gather}
where $w_s \in \mathbb{R}^{d}, W_s \in \mathbb{R}^{n_s \times d}, b_s \in \mathbb{R}^{n_s}$ are trainable parameters and $n_s$ is the number of possible sketches.

\subsection{Columns and Operators Prediction}
\label{sec:col-and-op}
To predict columns and operators, we use the LSTM decoder with the attention mechanism \citep{luong-attention} such that \textit{the number of decoding steps are decided by the sketch prediction module}. We train 5 different column prediction modules separately for each SQL clause, but they share the same architecture. 

In the column prediction module, the hidden state of the decoder at the $t$-th decoding step is computed as $d_{col}^{(t)} (\in \mathbb{R}^{d}) = \texttt{LSTM}(d_{col}^{(t-1)}, h_{col}^{(t-1)})$, where $h_{col}^{(t-1)} \in \mathbb{R}^{d}$ is an encoding of the predicted column in the previous decoding step. The context vector $r^{(t)}$ is computed by a weighted sum of question encodings $H_{Q} \in \mathbb{R}^{d \times \left\vert X \right\vert}$ based on attention weight as follows:
\begin{gather}
    \alpha^{(t)} = \texttt{softmax}({d_{col}^{(t)}}^T \, H_{Q}) \\
    r^{(t)} = H_Q \, {\alpha^{(t)}}^T
\end{gather}
Then, the attentional output of the $t$-th decoding step $a_{col}^{(t)}$ is computed as a linear combination of $d_{col}^{(t)} \in \mathbb{R}^{d}$ and $r^{(t)} \in \mathbb{R}^{d}$ followed by \texttt{tanh} activation.
\begin{equation}\label{eqn:attention}
    a_{col}^{(t)} = \texttt{tanh}(W_1 d_{col}^{(t)} + W_2 r^{(t)})
\end{equation}
where $W_1,W_2 \in \mathbb{R}^{d \times d}$ are trainable parameters.
Finally, the probability for each column at the $t$-th decoding step is computed as a dot product between $a_{col}^{(t)} \in \mathbb{R}^{d}$ and the encoding of each column in $H_{col} \in \mathbb{R}^{d \times \left\vert C \right\vert}$ followed by softmax.
\begin{equation}
    P_{col}^{(t)} = \texttt{softmax}({a_{col}^{(t)}}^T \, H_{col})
\end{equation}

To predict corresponding operators for each predicted column, we use a decoder of the same architecture as in the column prediction module. The only difference is that a decoder input at the $t$-th decoding step is an encoding of the $t$-th predicted column from the column prediction module.
\begin{equation}
    d_{op}^{(t)} = \texttt{LSTM}(d_{op}^{(t-1)}, h_{col}^{(t)})
\end{equation}
Attentional output $a_{op}^{(t)} \in \mathbb{R}^{d}$ is computed identically to Eq. (\ref{eqn:attention}). Then, the probability for operators corresponding to the $t$-th predicted column is computed by the softmax classifier as follows:
\begin{equation}
    P_{op}^{(t)} = \texttt{softmax}(W_o a_{op}^{(t)} + b_o)
\end{equation}
where $W_o \in \mathbb{R}^{n_o \times d}$ and $b_o \in \mathbb{R}^{n_o}$ are trainable parameters and $n_o$ is the number of possible operators.

\begin{table*}[h!]
\small
\centering
\begin{tabular}{l|ccccc|c}
\toprule
 & \multicolumn{5}{c}{Dev} & \multicolumn{1}{c}{Test} \\
Method & Easy & Medium & Hard & Extra Hard & All & All \\
\midrule
SQLNet & 23.2\% & 8.6\% & 9.8\% & 0\% & 10.9\% & 12.4\% \\
TypeSQL & 18.8\% & 5.5\% & 4.6\% & 2.4\% & 8.0\% & 8.2\% \\
SyntaxSQLNet & 38.4\% & 15.0\% & 16.1\% & 3.5\% & 19.0\% & 19.7\% \\
\midrule
Ours & \textbf{53.2\%} & \textbf{27.0\%} & \textbf{20.1\%} & \textbf{6.5\%} & \textbf{28.8\%} & \textbf{24.3\%} \\
-rec & 53.2\% & 27.0\% & 14.4\% & 2.9\% & 27.4\% & - \\
-rec - col-att & 46.4\% & 22.0\% & 12.1\% & 4.7\% & 23.4\% & - \\
-rec -col-att -sketch & 33.2\% & 18.6\% & 11.5\% & 4.7\% & 18.7\% & - \\
\bottomrule
\end{tabular}
\caption{\label{result-table-1} Accuracy of exact SQL matching with different hardness levels.}
\end{table*}

\begin{table*}[h!]
\small
\centering
\begin{tabular}{l|ccccc}
\toprule
Method & SELECT & WHERE & GROUP BY & ORDER BY & KEYWORDS \\
\midrule
SQLNet & 46.6\% & 20.6\% & 37.6\% & 49.2\% & 62.8\% \\
TypeSQL & 43.7\% & 14.8\% & 16.9\% & 52.1\% & 67.0\% \\
SyntaxSQLNet & 55.4\% & 22.2\% & 51.4\% & 50.6\% & 73.3\% \\
\midrule
Ours & \textbf{68.7\%} & \textbf{39.0\%} & \textbf{63.1\%} & \textbf{63.5\%} & \textbf{76.5\%} \\
\bottomrule
\end{tabular}
\caption{\label{result-table-2} F1 scores of SQL component matching on the \textit{dev} set. }
\end{table*}

\subsection{From Clause Prediction}
\label{sec:from}
After the predictions of all the other clauses, we use a heuristic to generate the \texttt{FROM} clause. We first collect all the columns that appear in the predicted SQL, and then we \texttt{JOIN} tables that include these predicted columns.


\subsection{Recursion for Nested Queries}
To predict the presence of a sub-query, we train another module that has the same architecture as the operator prediction module. Instead of predicting corresponding operators for each column, it predicts whether each column is compared to a variable (e.g., \texttt{WHERE} age $>$ 3) or to a sub-query (e.g., \texttt{WHERE} age $>$ (\texttt{SELECT} \texttt{avg}(age) ..)). In the latter case, we add the temporal \texttt{[SUB\_QUERY]} token to the corresponding location in the SQL output. Additionally, if the sketch prediction module predicts one of \texttt{INTERSECT/UNION/EXCEPT} operators, we add a \texttt{[SUB\_QUERY]} token after the operator.

To predict a sub-query, our model takes the temporal generated SQL with a \texttt{[SUB\_QUERY]} token as an input in addition to a natural language question with separate token \texttt{[SEP]} (e.g., What is ... \texttt{[SEP]} \texttt{SELECT} ... \texttt{INTERSECT} \texttt{[SUB\_QUERY]}). This input is encoded in the same way as question encoding described in Section~\ref{sec:encoding}. Then, the rest of the SQL generation process is identical to that described in Section~\ref{sec:sketch}--\ref{sec:from}. After the sub-query is predicted, it replaces the \texttt{[SUB\_QUERY]} token to form the final query.

\section{Experiments}

\subsection{Experimental Setup}
We evaluate our model with \textit{Spider} \citep{spider}, a large-scale, complex and cross-domain text-to-SQL dataset. We follow the same database split as \citet{spider}, which ensures that any database schema that appears in the training set does not appear in the dev or test set. Through this split, we examine how well our model can be generalized to unseen databases. Because the test set is not opened to the public, we use the \textit{dev} set for the ablation analysis.
For the evaluation metrics, we use 1) accuracy of exact SQL matching and 2) F1 score of SQL component matching, proposed by \citep{spider}. We also follow their query hardness criteria to understand the model performance on different levels of queries. Our model and all the baseline models are trained based on only the \textit{Spider} dataset without data augmentation.

\subsection{Model Configuration}
We use the same hyperparameters for every module. For the word embedding, we apply deep contextualized word representations (ELMO) from \citet{elmo} and allow them to be fine-tuned during the training. For the question and column encoders, we use a 1-layer 512-unit bi-directional LSTM. For the decoders in the columns and operators prediction modules, we use a 1-layer 1024-unit uni-directional LSTM. 
For the training, we use Adam optimizer \citep{adam} with a learning rate of 1e-4 and use early stopping with 50 epochs. Additionally, we use dropout \citep{dropout} with a rate of 0.2 for the regularization.

\subsection{Result and Analysis}
Table~\ref{result-table-1} shows the exact SQL matching accuracy of our model and previous models. We achieve 24.3\% and 28.8\% on the test and dev sets respectively, which outperforms the previous best model SyntaxSQLNet \citep{syntaxsqlnet} by 4.6\% and 9.8\%. Moreover, our model outperforms previous models on all different query hardness levels.

To examine how each technique contributes to the performance, we conduct an ablation analysis of three aspects: 1) without recursion, 2) without self-attention for database schema encoding, and 3) without sketch prediction modules that decide the number of decoding steps. Without recursive sub-query generation, the accuracy drops by 5.7\% and 3.6\% for hard and extra hard queries, respectively. This result shows that the recursion we use enables the model to predict nested queries. When using the final LSTM hidden state as in \citet{syntaxsqlnet} instead of using self-attention for schema encoding, the accuracy drops by 4.0\% on all queries. Finally, when using only an encoder-decoder architecture without sketch generation for columns prediction, the accuracy drops by 4.7\%.

For the component matching result for each SQL clause, our model outperforms previous approaches for all of the SQL components by a significant margin, as shown in Table~\ref{result-table-2}. Examples of predicted SQL from different models are shown in Appendix~\ref{sec:appendix-a}.

\section{Conclusion}
In this paper, we propose a recursive and SQL clause-wise decoding neural architecture to address the complex and cross-domain text-to-SQL task. We evaluate our model with the \textit{Spider} dataset, and the experimental result shows that our model significantly outperforms previous work for generating not only simple queries, but also complex and nested queries.

\section*{Acknowledgments}
We thank Yongsik Lee, Jaesik Yoon, and Donghun Lee (SAP) for their reviews and support. We also thank professor Sungroh Yoon, Jongyun Song, and Taeuk Kim (Seoul National University) for their insightful feedback and three anonymous reviewers for their helpful comments.

\bibliography{emnlp-ijcnlp-2019}
\bibliographystyle{acl_natbib}

\appendix


\section{Sample SQL Predictions}
\label{sec:appendix-a}
In Table~\ref{sample-sql}, we show some examples of predicted SQL queries from different models. We compare the result of our model with two of previous state-of-the-art models: SyntaxSQLNet \citep{syntaxsqlnet} and the modified version of SQLNet \citep{sqlnet} by \citet{spider} to support complex SQL queries.

\begin{table*}[hbt!]
\small
\centering
\begin{tabular}{llL{13cm}}
\toprule
hardness & type & description \\
\midrule
    easy & NL & What are the names of all the countries that became independent after 1950? \\
        \cline{2-3}
            & Truth & \texttt{SELECT} Name \texttt{FROM} country \texttt{WHERE} IndepYear $>$ 1950 \\
            \cline{2-3}
            & Ours & \texttt{SELECT} Name \texttt{FROM} country \texttt{WHERE} IndepYear $>$ ``[VAR]" \\
            \cline{2-3}
            & Syntax & \texttt{SELECT} Name \texttt{FROM} country \texttt{WHERE} GovernmentForm $=$ ``[VAR]" \\
            \cline{2-3}
            & SQLNet & \texttt{SELECT} T1.Name \texttt{FROM} city \texttt{as} T1 \texttt{JOIN} country \texttt{as} T2 \texttt{WHERE} T2.Population $>$ ``[VAR]" \\
\midrule
    medium & NL & Which city and country is the Alton airport at? \\
            \cline{2-3}
                & Truth & \texttt{SELECT} City, Country \texttt{FROM} airports \texttt{WHERE} AirportName $=$ ``Alton" \\
                \cline{2-3}
                & Ours & \texttt{SELECT} City, Country \texttt{FROM} airports \texttt{WHERE} AirportName $=$ ``[VAR]" \\
                \cline{2-3}
                & Syntax & \texttt{SELECT} Country, City \texttt{FROM} airports \texttt{WHERE} Country $=$ ``[VAR]" \\
                \cline{2-3}
                & SQLNet & \texttt{SELECT} T1.City, T2.DestAirport \texttt{FROM} airports \texttt{as} T1 \texttt{JOIN} flights \texttt{as} T2 \\
\midrule
    medium & NL & List the names of poker players ordered by the final tables made in ascending order. \\
            \cline{2-3}
                & Truth & \texttt{SELECT} T1.Name \texttt{FROM} people \texttt{as} T1 \texttt{JOIN} poker\_player \texttt{as} T2 \texttt{ORDER BY} T2.Final\_Table\_Made \\
                \cline{2-3}
                & Ours & \texttt{SELECT} T1.Name \texttt{FROM} people \texttt{as} T1 \texttt{JOIN} poker\_player \texttt{as} T2 \texttt{ORDER BY} T2.Final\_Table\_Made \texttt{ASC} \\
                \cline{2-3}
                & Syntax & \texttt{SELECT} T2.Name \texttt{FROM} poker\_player \texttt{as} T1 \texttt{JOIN} people \texttt{as} T2 \texttt{ORDER BY} T1.Earnings \texttt{ASC} \\
                \cline{2-3}
                & SQLNet & \texttt{SELECT} Name \texttt{FROM} people \texttt{ORDER BY} Birth\_Date \texttt{ASC} \\
\midrule
    medium & NL & How much does the most recent treatment cost? \\
        \cline{2-3} & Truth & \texttt{SELECT} cost\_of\_treatment \texttt{FROM} Treatments \texttt{ORDER BY} date\_of\_treatment \texttt{DESC LIMIT} 1\\
            \cline{2-3}
            & Ours & \texttt{SELECT} cost\_of\_treatment \texttt{FROM} Treatments \texttt{ORDER BY} cost\_of\_treatment \texttt{DESC LIMIT} ``[VAR]"\\
            \cline{2-3}
            & Syntax & \texttt{SELECT} cost\_of\_treatment \texttt{FROM} Treatments \texttt{ORDER BY} cost\_of\_treatment \texttt{ASC LIMIT} ``[VAR]" \\
            \cline{2-3}
            & SQLNet & \texttt{SELECT} T1.charge\_amount \texttt{FROM} Charges \texttt{as} T1 \texttt{JOIN} Dogs \texttt{as} T2 \texttt{ORDER BY} date\_adopted \texttt{DESC LIMIT} ``[VAR]" \\
\midrule
    hard & NL & List the names of teachers who have not been arranged to teach courses. \\
            \cline{2-3}
                & Truth & \texttt{SELECT} Name \texttt{FROM} teacher \texttt{WHERE} Teacher\_id \texttt{NOT IN} (\texttt{SELECT} Teacher\_id \texttt{FROM} course\_arrange) \\
                \cline{2-3}
                & Ours & \texttt{SELECT} Name \texttt{FROM} teacher \texttt{WHERE} Teacher\_id \texttt{NOT IN} (\texttt{SELECT} Teacher\_id \texttt{FROM} course\_arrange) \\
                \cline{2-3}
                & Syntax & \texttt{SELECT} Name \texttt{FROM} teacher \\
                \cline{2-3}
                & SQLNet & \texttt{SELECT} Name \texttt{FROM} teacher \\
\midrule
    hard & NL & Which cities do more than one employee under age 30 come from? \\
        \cline{2-3}
            & Truth & \texttt{SELECT} city \texttt{FROM} employee \texttt{WHERE} age $<$ 30 \texttt{GROUP BY} city \texttt{HAVING} count(*) $>$ 1  \\
        \cline{2-3}
            & Ours & \texttt{SELECT} city \texttt{FROM} employee \texttt{WHERE} age $<$ ``[VAR]" \texttt{GROUP BY} city \texttt{HAVING} count(*) $>$ ``[VAR]" \\
        \cline{2-3}
            & Syntax & \texttt{SELECT} city \texttt{FROM} employee \texttt{WHERE} age $>$ ``[VAR]" \\
        \cline{2-3}
            & SQLNet & \texttt{SELECT} T1.city \texttt{FROM} employee \texttt{as} T1 \texttt{JOIN} hiring \texttt{as} T2 \texttt{JOIN} shop \texttt{as} T3 \texttt{WHERE} T3.District $>$ ``[VAR]" \texttt{GROUP BY} T1.city \texttt{HAVING} count(*) $>$ ``[VAR]" \\
\midrule
    hard & NL & What is the document id with least number of paragraphs? \\
        \cline{2-3} & Truth & \texttt{SELECT} document\_id \texttt{FROM} Paragraphs \texttt{GROUP BY} document\_id \texttt{ORDER BY} count(*) \texttt{LIMIT} 1 \\
            \cline{2-3}
            & Ours & \texttt{SELECT} document\_id \texttt{FROM} Documents \texttt{GROUP BY} document\_id \texttt{ORDER BY} count(*) \texttt{ASC LIMIT} ``[VAR]" \\
            \cline{2-3}
            & Syntax & \texttt{SELECT} document\_id \texttt{FROM} Documents \texttt{GROUP BY} document\_id \texttt{ORDER BY} count(*) \texttt{ASC LIMIT} ``[VAR]" \texttt{HAVING} count(*) $>=$ ``[VAR]"\\
            \cline{2-3}
            & SQLNet & \texttt{SELECT} template\_id \texttt{FROM} Templates \texttt{GROUP BY} template\_id \texttt{HAVING} sum(*) \texttt{NOT} ``[VAR]" \texttt{ORDER BY} count(*) \texttt{ASC LIMIT} ``[VAR]"\\
\midrule
    extra & NL & How many dogs have not gone through any treatment? \\
            \cline{2-3}
                & Truth & \texttt{SELECT} count(*) \texttt{FROM} Dogs \texttt{WHERE} dog\_id \texttt{NOT IN} (\texttt{SELECT} dog\_id \texttt{FROM} Treatments) \\
                \cline{2-3}
                & Ours & \texttt{SELECT} count(*) \texttt{FROM} Dogs \texttt{WHERE} dog\_id \texttt{NOT IN} (\texttt{SELECT} dog\_id \texttt{FROM} Treatments) \\
                \cline{2-3}
                & Syntax & \texttt{SELECT} count(*) \texttt{FROM} Charges \texttt{WHERE} charge\_id \texttt{NOT IN} (\texttt{SELECT} charge\_id \texttt{FROM} Charges) \\
                \cline{2-3}
                & SQLNet & \texttt{SELECT} count(*) \texttt{FROM} Dogs \texttt{WHERE} dog\_id \texttt{IN} ``[VAR]" \\
\midrule
    extra & NL & What is the name of the high schooler who has the greatest number of friends? \\
            \cline{2-3}
                & Truth & \texttt{SELECT} T2.name \texttt{FROM} Friend \texttt{as} T1 \texttt{JOIN} Highschooler \texttt{as} T2 \texttt{GROUP BY} T1.student\_id \texttt{ORDER BY} count(*) \texttt{DESC} \texttt{LIMIT} 1 \\
                \cline{2-3}
                & Ours & \texttt{SELECT} T1.name \texttt{FROM} Highschooler \texttt{as} T1 \texttt{JOIN} Friend \texttt{as} T2 \texttt{GROUP BY} T2.student\_id \texttt{ORDER BY} count(*) \texttt{DESC} \texttt{LIMIT} 1 \\
                \cline{2-3}
                & Syntax & \texttt{SELECT} name \texttt{FROM} Highschooler \texttt{ORDER BY} grade \texttt{DESC} \texttt{LIMIT} 1 \\
                \cline{2-3}
                & SQLNet & \texttt{SELECT} T1.name \texttt{FROM} Friend \texttt{as} T1 \texttt{JOIN} Friend \texttt{as} T2 \texttt{GROUP BY} T2.student\_id \texttt{ORDER BY} * \texttt{DESC} \texttt{LIMIT} 1 \\
\bottomrule
\end{tabular}

\caption{\label{sample-sql} Sample SQL predictions by our model and previous state-of-the-art models on the dev split. \texttt{NL} denotes the natural language question and \texttt{Truth} denotes the corresponding ground truth SQL query. \texttt{Ours}, \texttt{Syntax}, and \texttt{SQLNet} denotes the SQL predictions from our model, SyntaxSQLNet \citep{syntaxsqlnet}, and modified SQLNet \citep{sqlnet} by \citet{spider}, respectively. }
\end{table*}

\end{document}